\documentclass[11pt,a4paper]{article}
\usepackage[hyperref]{eacl2021}
\usepackage{times}
\usepackage{latexsym}

\usepackage{microtype}
\usepackage{multirow}
\usepackage{graphicx} 
\aclfinalcopy 
\def\aclpaperid{336} 

\title{Continuous Learning in Neural Machine Translation using Bilingual Dictionaries}

\author{Jan Niehues \\
  Department of Data Science and Knowledge Engineering (DKE) \\
  Maastricht University, Maastricht, The Netherlands \\
  \texttt{jan.niehues@maastrichtuniversity.nl} }

\date{}

\begin{document}
\maketitle
\begin{abstract}

While recent advances in deep learning led to significant improvements in machine translation, neural machine translation is often still not able to continuously adapt to the environment. 
For humans, as well as for machine translation, bilingual dictionaries are a promising knowledge source to continuously integrate new knowledge. However, their exploitation poses several challenges: The system needs to be able to perform one-shot learning as well as model the morphology of source and target language.

In this work, we proposed an evaluation framework to assess the ability of neural machine translation to continuously learn new phrases. We integrate one-shot learning methods for neural machine translation with different word representations and show that it is important to address both in order to successfully make use of bilingual dictionaries. By addressing both challenges we are able to improve the ability to translate new, rare words and phrases from 30\% to up to 70\%. The correct lemma is even generated by more than 90\%.

\end{abstract}

\section{Introduction}

Recent advances in neural machine translation (NMT) have led to astonishing translation quality of research systems in evaluation campaigns as well as for commercial systems. These improvements even led to discussions whether automatic machine translation is already on par with human translation \citep{barrault_findings_2019}.
One challenge that has raised less attention is the ability of these systems to continuously learn over time. In contrast, humans are continuously improving their skills and adapting to an ever-changing environment.

\begin{table*}[ht]
 \begin{center}
   \begin{tabular}{ll}
Source: &  Tell us, what have you got against \textbf{giraffes} ?\\
Dictionary: & giraffe $\rightarrow$ Giraffe \\
Reference: & was haben Sie eigentlich gegen \textbf{Giraffen} ? \\
Annotation: & Tell us, what have you got against \# giraffes \# Giraffe \# ? \\
\end{tabular}   
\end{center}
\caption{\label{Example} Example of dictionary usage }

\end{table*}

There are several reasons why this is necessary: First, nobody is fluent in all possible domains. Even professional translators need to adapt to the specific vocabulary of different domains. Secondly, language is not static but developing over time and translators need to learn new terms, meanings and expressions.

For humans, one successful approach to adapt to the environment is the usage of a dictionary \footnote{In this work the dictionary entries can consist of a single word or whole phrases}. Learning translations from a dictionary has several advantages: Dictionaries contain minimal examples. We do not need to collect full sentences, but can directly learn translations from a single phrase. Furthermore, this can even be generalized to other inflected forms of the same lexem. 
Secondly, it enables the system to directly integrate correction. If a user sees a specific problem, the user can interact with the system by adding a specific dictionary entry. This is very important if a specific terminology should be used. 

Motivated by the success for human translators, in this work we will enable NMT to also successfully integrate knowledge from bilingual dictionaries. Thereby, we will focus on learning translations that could not be learned from parallel data. This poses several interesting research challenges as shown in the example in Table \ref{Example}. When training a system on the proceedings of the European Parliament, it might never have seen the word \textit{giraffe} and needs to learn the translation from the dictionary. First of all, we have to address one-shot learning. The system needs to be able to continuously learn new dictionary entries and then should directly be able to translate all occurrences of this phrase.

Secondly, the model must be aware of the morphology of the source and target language. In a dictionary only the base form of a word is given. In the example only the lemma \textit{giraffe} is in the dictionary, but not the plural form \textit{giraffes}. Therefore, we must enable the system to translate different lexemes of a lemma by knowing only the translation of the base form. This involves analysing the morphological form of the source word, transferring the information about the form to the target and finally generating the correct morphological form of the target word based on the dictionary entry as well as on the morphological form of the source word. In German the plural of the dictionary entry \textit{Giraffe} is \textit{Giraffen}.

In order to assess the approaches on this challenging condition, it is essential to define an appropriate evaluation scheme. While the ability to continuously learn new translations is essential in many practical applications, the newly learned terminology will only occur rarely. Therefore, standard methods for evaluating machine translation are not able to measure the effect appropriately.

In order to address these challenges, we develop the following contributions:
\begin{itemize}
    \item We developed a targeted evaluation approach for the continuous learning of new translations (Section \ref{Evaluation}) 
    \item We showed that character-based representation is essential to inflect unknown words correctly. (Section \ref{Model})
    \item We show that only the combination of word representation and one-shot learning enables the successful integration of bilingual dictionaries (Section \ref{Model})
\end{itemize}

\section{Evaluation scenario}
\label{Evaluation}
The first important research question that needs to be addressed in the targeted continuous learning scenario is the evaluation approach. While the evaluation of machine translation is well-established (e.g. using BLEU \citep{papineni-etal-2002-bleu}), new learned words are typically rare words and therefore their influence on a BLEU score calculated on all words is very limited.

In order to have a valid evaluation approach, the evaluation should focus on phrases that cannot be learned from the parallel data. These are typically very rare phrases. Furthermore, we want to translate them in a real world situation. Therefore, the evaluation data should not be synthetic sentences. Finally, the approach should be using the standard parallel data without the need of collecting additional parallel data.

A first attempt would be to use existing test data and select sentences where dictionary entries are needed as e.g. done in \cite{dinu_training_2019}. However, if we limit ourselves to phrases that do not occur in the parallel data or only a few times, the number of occurring words in the test sets are too low to draw any conclusions.

\begin{figure*}[h] 
\centering
         \includegraphics[width=.75\textwidth]{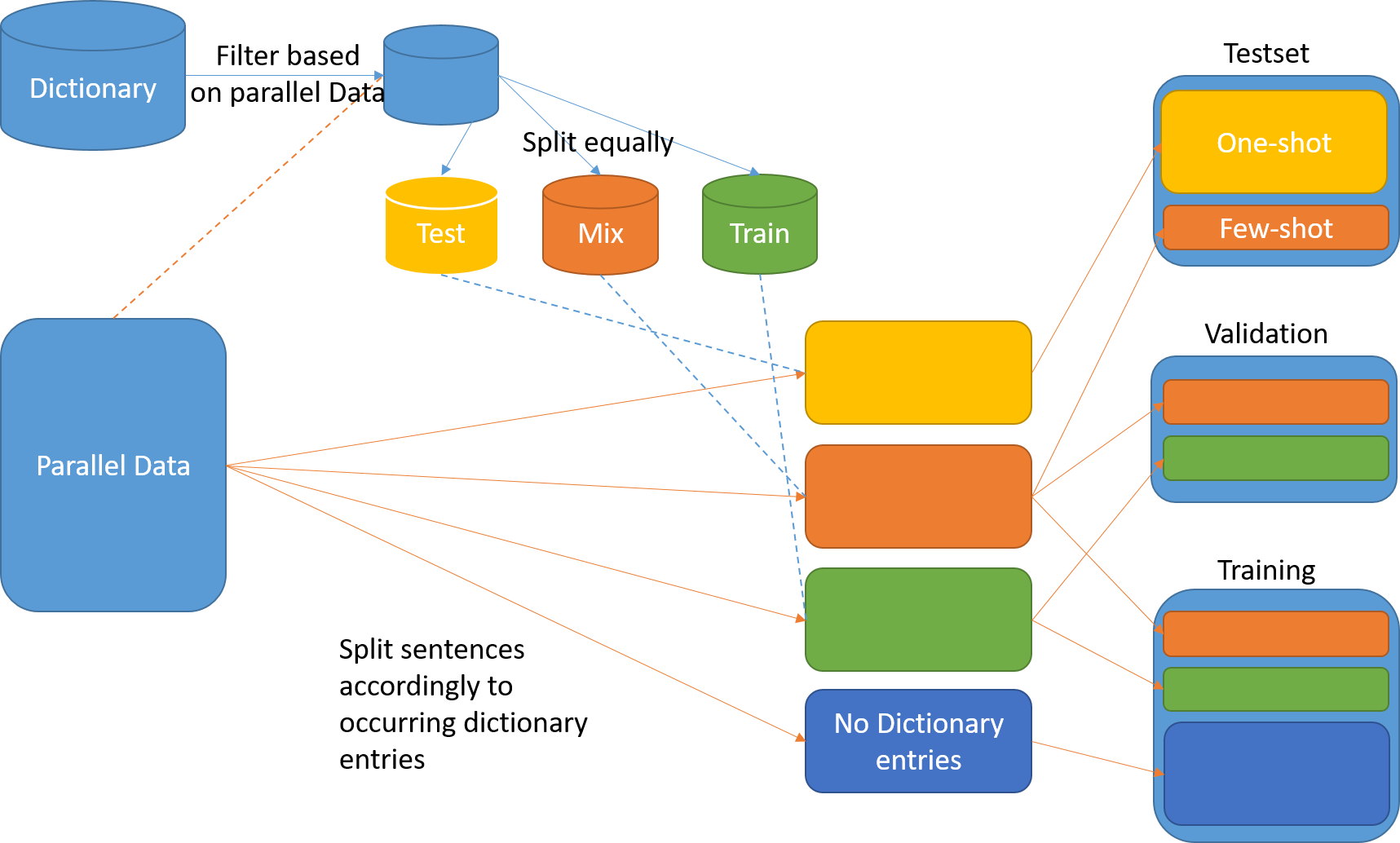}
\caption{Overview of the evaluation approach: Based on the parallel data and dictionary, a new split of the data is generated}
\label{fig:eval}
\end{figure*}

Therefore, we evaluate our approach by proposing a new test-train split of existing parallel data. In a first step, we filter a large background dictionary for entries that help to translate phrases that only occur a few times in the existing parallel data. In a second step, we select some of the sentences with their matching dictionaries entries as the new test sets. An overview of the process is shown in Figure \ref{fig:eval}. Finally, we specifically evaluate the ability of the translation system to translate the dictionary entries.

In addition, it is important to ensure that the proposed methods do not have negative side effects on the overall translation quality. Therefore, we also evaluate the model using standard evaluation metrics on well-established test sets and on the proposed test set. Due to the weakness of these metrics to measure improvements in rare words, we do not expect that the proposed methods improve on these metrics, but it is important that the performance measured in these metrics does not decrease significantly.

\subsection{Dictionary filtering}

In a first step, we create a large background dictionary for each considered language pair by extracting a bilingual dictionary from the English Wiktionary. Therefore, we extracted the translation from a Wiktionary dump\footnote{ https://dumps.wikimedia.org/enwiktionary/20200501/enwiktionary-20200501-pages-articles.xml.bz2} using wiktextract \footnote{https://github.com/tatuylonen/wiktextract}.

Secondly, we match the dictionary entries to the targeted corpus. Therefore, we lemmatize all dictionary entries as well as both sides of the parallel data. This is done to also find matches for all morphological variants of the dictionary entries. Finally, we calculate the statistics mentioned in Table \ref{Stat} for each dictionary entry about its matches to the parallel data.

In a third step, we filter the dictionary based on the statistics. We only select words that are rare in the corpus. If the words are common and occur often in the training data, a dictionary entry would not be helpful. Secondly, we want to analyse the ability of the system to generate different morphological forms. Therefore, we only consider entries that occur at least with two different morphological variants on the target side. Finally, in this work we focus on words that are not ambiguous. We leave an integration of word sense disambiguation to also handle ambiguous dictionary entries for future work. Therefore, we only consider phrases, where both, the source and target phrase, occur less than $10$ times with a different translation than the one given in the dictionary. 

\begin{table}[ht]
 \begin{center}
   \begin{tabular}{ll} \hline
   Statistic & Threshold \\ \hline
Occurrences & $ 3 \le k \le 80 $ \\ 
target inflected phrases & $\ge 2 $ \\
only source/target match & $< 10$ \\ \hline
\end{tabular}   
\end{center}
\caption{\label{Stat} Dictionary filtering }

\end{table}

\subsection{Train-Test Split}

Finally we generate a split of the corpus into training, validation and test sets based on the selected dictionary entries as shown in Figure \ref{fig:eval}. The model needs to learn how to use the dictionary. Therefore, several training sentences need to be annotated with dictionary entries. Furthermore, we want to evaluate the ability of the model to translate phrases it has seen a few times in training (\textit{Few-Shot learning}) as well as words it only has seen in the dictionary (\textit{One-shot learning}). 
Therefore, we split the entries in the dictionary equally into three sets (\textit{Test (yellow)}, \textit{Mix (orange)} and \textit{Train (green)}). All sentence pairs associated with entries from the \textit{Test} set are added to the newly created test set. 

In a second step, we select all the sentences from the remaining training sentences, where an entry from the \textit{Mix} set occurs. For each entry, half the sentences are added to the test set and a quarter to the validation and training set.

Finally, all sentences with entries from \textit{Train} are equally distributed to the training and validation set. Since we want to concentrate on modelling the morphology when using the dictionary and not the translation ambiguity for dictionary entries, we removed all sentences from the training where the source entry from the dictionary occurs, but the target sentence does not contain the target entry. Due to our selection of the dictionary, where we focus on words that have only very few different translations (less than $10$ times a different one), we only removed very few sentences here. All remaining sentences with no annotations (most of the sentences) were used for training.

\subsection{Evaluation}
\label{Ana}

When evaluating we want to focus on the system's ability to translate the phrases from the dictionary. Therefore, we measure the accuracy of translating the dictionary entries in addition to the commonly used BLEU score. In addition to calculating the accuracy by comparing the inflected words of hypothesis and reference (\textit{Exact match}), we calculate further statistics to analyse the approaches.

In addition, we measure the ability of the system to at least create the correct lemma by ignoring errors made due to wrong inflection of the words. Therefore, for each sentence, we compare the target lemmatized phrase of the dictionary entry with the lemmatized version of the generated translation. We will refer to this metric as \textit{Lemma match}

Finally, we are especially interested in the ability of the model to generate the correct inflected form. For many words, this is quite straightforward since it is the same as the lemma. Therefore, we also measure the exact match on the subset of the dictionary entries, where the target side of the dictionary is different from the inflect form occurring in the reference. To generate the correct translation, in this case the model really needs to change the output. We will refer to this metric as \textit{Morph. Adjustment}.

In addition to these three evaluation scores, we also investigate the performance on the different types of entries. We evaluate all metrics on all entries and independently on the one-shot (\textit{OneS}) and few-shot (\textit{FewS}) entries.

\section{NMT Dictionary Integration}
\label{Model}

To successfully integrate the dictionary into the NMT system, we need to address two challenges: First, we need to enable the system to perform one-shot learning. It should be able to translate a phrase after seeing it only once in the bilingual dictionary. Furthermore, it needs to be possible to continuously add new translations. Secondly, we need to model the morphology of the dictionary entries. We need to use the dictionary for different inflected forms of the word and also generate various inflected forms of the target phrase.

\subsection{One-shot learning}
\label{OneShot}

In order to achieve one-shot learning, we need to combine the dictionary with our neural machine translation system. The combination should ensure fast learning, so a single dictionary entry is enough to learn the translation. Furthermore, it needs to be flexible, so new dictionary entries can be continuously added to the system and it is able to perform life-long learning by using the newly added entries.

One large advantage of deep learning approaches is that they are able to easily incorporate additional information. By annotating the input with additional information, the model is able to learn automatically how to make use of this additional information. This has been successfully done, for example, for the translation of other MT systems \citep{niehues_pre-translation_2016}, for domain information \citep{kobus_domain_2017} or information about formality \citep{sennrich_controlling_2016}.

For the integration of additional knowledge about specific phrases, we follow similar approaches presented in \citet{pham_towards_2018} and \citet{dinu_training_2019}. The main idea is that we annotate each source phrase, for which a dictionary translation is available with this translation. This is done by appending the translation to the source phrase within the sentence as shown in Table \ref{Example}. Since this is done during training and testing, the system is able to learn to copy and modify these suggestions. No further adaptation to the architecture of the NMT system is necessary. The system will learn how to exploit these systems and can transfer this knowledge to new translations that have not been seen in training. Therefore, the translations need only to be added once to the dictionary, which enables the system to perform one-shot learning as well as to continuously learn new translation by extending the dictionary.

The main difference to previous work is that we are focusing on very rare words and morphological variants of the dictionary phrases. Therefore, we investigate the matching of the dictionary entries as well as the number of necessary entries.

In order to find the dictionary entries for a given source sentence, we first lemmatize the sentence. In a second step, we then match the dictionary to the lemmatized sentences. Finally, we map back the found entries to the original sentence.

In annotating the source sentence, we follow the related work and append the translation to the source phrase. As shown in Figure \ref{Example}, we replace the source word \textit{giraffes} by the entry \textit{\# giraffes \# Giraffe \#}. In contrast to the original work, we do not have the inflect target words, but only put the lemmatized target string to the sentence. For the source side, we keep the inflected form for the source sentence so the system is able to extract important morphological information from the source (e.g. grammatical number) and map it to the target. This is done for the training and test data. Then the baseline neural machine translation system is trained normally on the annotated sentences. We did not adapt the architecture since in \citet{dinu_training_2019} the standard transformer based system was able to learn to copy the suggested translations into the target side.

While the system should learn to also use dictionary entries it has not seen during training, the system needs enough examples in order to learn how to use dictionary entries in general. Since we are concentrating on very rare words, the number of dictionary entries in the parallel data is relatively small. For larger corpora, we therefore explore whether it is helpful to annotate additional phrases. This was done by also extracting phrases that occur more often (\textit{add. Annot}). However, we did use the same split and also evaluated our approach only on the rare phrases.

\subsection{Word representations}
\label{Representation}
A second challenge when building a machine translation system for the targeted scenario is the generation of the correct inflected word form. Since we have seen the new words only in the dictionary, we will often need to generate different inflected word forms that we have neither seen in the dictionary nor in the corpus.

While there have been attempts to generate unknown inflected word forms for dictionary entries (e.g. \citet{niehues_using_2011}) prior to neural machine translation, the ability to represent parts of the words in neural machine translation offer a unique opportunity to model morphological inflection. Therefore, in this work, we concentrated on the word representation used in the NMT system. Thereby, we always use the same representation for the source and the target language. The most commonly used word representation used in state-of-the-art neural machine translation systems are byte-pair-encodings (BPE) \citep{sennrich_neural_2016}. A second successful approach to represent words in a neural machine translation system are character-based representations, where each word is split into its characters.

While there have been several works on comparing these two representations(e.g. \citet{sennrich_how_2017}, they are mostly concentrating on generating the overall best translation performance. However, in this work, we will focus on the rare words. Since only for these words we need to learn how to generate different inflected forms. For the more frequent words, this is often not that important since all word forms occur several times in the corpus.

Besides the generation of unknown inflected forms, the word representation is also important when learning to copy the annotations to the target. If we look at the example dictionary entry \textit{ concentric $\rightarrow$ konzentrisch}, the lemma \textit{konzentrisch} got split into the subwords \textit{konzent@@ ris@@ ch} while the inflect form \textit{konzentrischer} into \textit{kon@@ zentr@@ ischer}. In this case there is no overlap in the subwords between the lemma and the inflected form. Therefore, it is difficult for the system to learn from the suggested translation. In contrast, when looking at the character-based representation, the model can copy the lemma and only has to learn to add additional tokens at the end.

In a first step, we compared character-based and sub-word based models. Thereby, we highlight their ability to generate new inflected forms of rare words. For both we used exactly the same NMT architecture. The only difference is that the input and output length for the character-based models is significantly larger since the number of characters is higher than the number of subwords.

We will see that the character-based models are significantly better in generating the different inflected forms for rare words. However, a major challenge is the training time. Due to the significant longer sequence length, also the training and decoding time is much slower. Therefore, we also propose a combination of word-based and character-based models.

In the mixed representation, we split each word that occurs less than $k$ times into its characters, while the other words are kept as they are. Since only frequent words are not split into characters, no further subword segmentation for these words is performed. Thereby, we can speed up the processing due to a short sequence length, but still have the ability to learn how to inflect rare words. Some dictionary entries contain phrases with many frequent words. In order to be able to better inflect these words, in a second approach we in addition split also all words within a dictionary phrase into characters. We refer to this technique as \textit{Mix+Annot}.

\section{Experiments}

We evaluate the approaches on three different data sizes and on two different language pairs (English-German and English-Czech). Since we are focusing on the generation of different morphological forms, we always use the morphologically rich language as the target language.

\subsection{Data}

For English-to-German we created two datasets with different sizes. A first series of experiments is run on the TED \citep{cettolo_wit_2012} corpus. We split the corpus into training, validation and test sets as described in Section \ref{Evaluation}. In addition, we evaluate the system also on the official test sets \textit{tst2014}, \textit{tst2015} and \textit{2018} and report average metrics for these test sets.

For the second system, we use the Europarl corpus \citep{koehn_europarl_2005}. This corpus is around 10 times bigger than the TED corpus as shown in Table \ref{SIZE}. In addition to the target test set, we also tested the systems on the test2006 and test2007, which are the most recent official test sets from the same domain used for the WMT.

Finally, we also tested the techniques on a different language pair. For this we choose English to Czech and also use the Europarl corpus for these experiments. Since there is no official in-domain corpus available, we tested the systems also on the newstest2019 test set.

As shown in Table \ref{SIZE}, the parameters mentioned in Section \ref{Evaluation} lead to a reasonable test set size for all corpora. As mentioned in Section \ref{OneShot}, we evaluate the system on Europarl with different amounts of training annotations. All data sets with their splits are available for further experiments \footnote{https://nlp-dke.github.io/data/rareWordNMT/}.

\begin{table}[ht]
 \begin{center}
   \begin{tabular}{lrrr} \hline 
& \multicolumn{2}{c}{EN-DE} & EN-CS \\
& TED & Europarl & Europarl \\ \hline
Train & 198K & 1.9M & 636K \\
 - Annot & 1.6K & 1.2K & 2.7K \\
 - add. Annot & & 14.5K & 24.3K \\
Valid & 1610 & 1196 & 2000 \\
Test & 3181 &  2140 & 5360 \\ \hline
\end{tabular}   
\end{center}
\caption{\label{SIZE} Data size in number of sentences }
\end{table}

\subsection{System}

All data was processed using the Stanza toolkit \citep{qi_stanza_2020} for tokenization and lemmatization. The lemmatization was only used for matching the dictionary entries, the translation systems were built on the inflected words. If BPE is applied, we used a BPE size of 20K. For the mixed representation, words occurring less than $k=50$ times were represented as individual characters.

We use the standard transformer architecture \citep{vaswani_attention_2017} and increase the number of layers to eight. The layer size is 512 and the inner size is 2048. Furthermore, we apply word dropout \citep{gal_theoretically_2016} with $p=0.1$.  We use the same learning rate schedule as in the original work and the implementation presented in \cite{pham_very_2019} \footnote{https://github.com/nlp-dke/NMTGMinor}. All systems were always trained from scratch with random initialization.

\subsection{TED}
\begin{table*}[!ht]
 \begin{center}
   \begin{tabular}{llcccc} \hline 
\multirow{2}{*}{Represen-tation} & \multirow{2}{*}{One-Shot} &  \multicolumn{2}{c}{ CL Test} & \multicolumn{2}{c}{official Test} \\ 
&& BLEU $\uparrow$ & characTER $\downarrow$ & BLEU $\uparrow$ & characTER $\downarrow$ \\ \hline
BPE & No & 25.97 & 44.09 & 26.17 & 44.62\\
Character & No & 28.12 & 42.79 & 26.57 & 44.27\\ 
Mix & No & 27.44 & 42.79 & 26.83 & 44.28\\
BPE & Annot & 26.00 & 41.74 & 26.21 & 44.73\\
Character & Annot & 28.92 & 40.16 & 26.72 & 43.96\\ 
Mix & Annot & 28.93 & 40.96  & 26.8 & 44.44\\ \hline

\end{tabular}   
\end{center}
\caption{\label{TEDTranslationMetric} Translation quality on TED tasks }
\end{table*}

\begin{table*}[!ht]
 \begin{center}
   \begin{tabular}{ll|ccc|ccc|ccc} \hline 
\multirow{2}{*}{Representation} & \multirow{2}{*}{One-Shot} &  \multicolumn{3}{|c|}{Exact match} & \multicolumn{3}{|c|}{Lemma match} & \multicolumn{3}{|c}{Morph. Adjustment} \\ 
&& All & OneS & FewS & All & OneS & FewS & All & OneS & FewS \\ \hline
BPE & No & 34 & 22 & 53 & 31 & 27 & 62 & 29 & 22 & 43\\
Character & No &  48 & 40 & 60 & 55 & 47 & 68 & 45 & 43 & 48\\ 
Mix & No & 42 & 35 & 54 & 49 & 40 & 63 & 38 & 34 & 46\\ \hline
BPE & Annot & 48 & 34 & 69 & 62 & 46 & 88 & 33 & 24 & 50\\
Character & Annot & 76 & 74 & 78 & 92 & 91 & 93 & 62 & 61 & 64\\ 
Mix & Annot & 75 & 72 & 79 & 92 & 91 & 94 & 59 & 56 & 65\\ \hline

\end{tabular}   
\end{center}
\caption{\label{TEDTranslationAna} Rare word accuracy on TED tasks }
\end{table*}

A first series of experiments were performed on the TED task. We evaluated the one-shot learning approach by source sentence annotation as well as the three different word representations described in Section \ref{Representation}. In a first step, we evaluated the translation performance using BLEU (mteval-v14.pl) and characTER \citep{wang_character:_2016} on the continuous learning test set as well as on the official test set (Table \ref{TEDTranslationMetric}).

The baseline systems using no one-shot learning do not annotate the source at all and are trained on the standard parallel data. If we take a look at the official test set, we see systems using character-based representation (\textit{Character} and \textit{Mix}) perform slightly better than the subword-based models. This might be due to the fact that the TED training data is rather small. Secondly, the one-shot learning approach has no influence on the translation performance of this test set. This is not surprising, since only 94 phrases in the 4343 sentences of the test sets were annotated. Therefore, we also evaluated our approach on the dedicated continuous-learning test set (\textit{CL test}), created by the new train-test split.

The improvements by character-based representation on the CL test set are even larger. This might be due to the fact that there are more rare words in these sentences and therefore the advantages of the character-based models is stronger. Secondly, in this case, the one-shot approach improvements improve the translation quality. Since the improvements for the BPE-based system are only measured by characTER and not by BLEU might indicate that for this system it is more challenging to generate the correct inflected form. 

To better analyse this, we also perform a detailed evaluation as described in Section \ref{Ana} and shown in Table \ref{TEDTranslationAna}. 
First of all, the experiments show the difficulty of the task. The baseline system is only able to translate 34\% of the phrases correctly. For the one-shot subset this even drops to 22\%.

Secondly, the experiments show that the challenge can only successfully be addressed by modelling both: one-shot learning and word representation. On the last two lines using character-based word representation and one-shot learning are able to achieve high accuracy. We see an improvement by 50\% percent absolutely, which is a relative improvement by more than 300\%. Furthermore, for these models there is no longer a clear difference between the one-shot and few-shot examples (Comparison of Columns \textit{OneS} and \textit{FewS}).

By looking at them separately, we see that only using one-shot learning improves the quality slightly. However, even when ignoring the word infection, the model often is not able to produce the correct lemma. The example in Section \ref{Representation}, motivates one challenge when learning to copy with different subword segmentations. If we only use character-based representations, we see improvements, especially for phrases that do not occur in training. In this case, the model is more often able to find the correct translation based on translations of other words. However, a similar performance between the few-shot and one-shot learning is only achieved by combining both techniques.

Finally, when only looking at the words where the lemma is different from the inflected form, we still see open research challenges. While we also could improve the accuracy from around 20\% or 30\% to nearly 60\%, it is still the most difficult case. 

While there is no clear difference between the character-based model and the mixed model on the output quality, there is a clear difference in training speed. For the full training on 64 epochs, the character-based model needs 14h, while the mixed representation only needs around 4h. While this is still slower than the subword-based model (2.5h), it still allows for a fast training of the model. Therefore, we only compared the mixed and the sub-word based representation for the remaining experiments on larger corpora.

\subsection{Europarl}

\begin{table*}[!ht]
 \begin{center}
   \begin{tabular}{lllcccc} \hline 
\multirow{2}{*}{Lang.}  & Represen- & \multirow{2}{*}{One-Shot} &  \multicolumn{2}{c}{ CL Test} & \multicolumn{2}{c}{official Test} \\ 
& tation && BLEU $\uparrow$ & characTER $\downarrow$ & BLEU $\uparrow$ & characTER $\downarrow$ \\ \hline
\multirow{7}{*}{Ger.} & BPE & No & 28.74 & 47.30 & 25.30 & 48.75\\
&Mix & No & 30.83  & 45.80 & 25.52 & 48.48\\ \cline{2-7}
&BPE & Annot & 28.74  & 47.47 & 25.49 & 48.64\\
&Mix & Annot & 31.63 & 44.64 & 25.45 & 48.60\\ \cline{2-7}
&BPE & add. Annot & 28.81 & 47.24 & 25.45 & 48.71\\
&Mix & add. Annot & 31.44 & 44.75 & 25.50 & 48.57\\ 
&Mix+Annot & add. Annot &  31.76 & 44.11 & 25.64 & 48.41\\ \hline
\multirow{5}{*}{Czech} & BPE & No & 34.25 & 39.43 & 16.2 & 57.15\\
 & BPE &  Annot & 34.73 & 38.39 & 15.57 & 57.59\\
 & Mix+Annot & Annot & 34.86 & 38.16 & 16.62 & 57.70\\ \cline{2-7}
 & BPE & add. Annot & 34.74 & 38.89 & 15.7 & 57.37\\
 & Mix+Annot & add. Annot & 35.21 & 37.95 & 16.63 & 57.65\\ \hline

\end{tabular}   
   \begin{tabular}{lll|ccc|ccc|ccc} \hline 
\multirow{2}{*}{Lang.}  & Represen- & \multirow{2}{*}{One-Shot} &  \multicolumn{3}{|c|}{Exact match} & \multicolumn{3}{|c|}{Lemma match} & \multicolumn{3}{|c}{Morph. Adjustment} \\ 
& tation && All & OneS & FewS & All & OneS & FewS & All & OneS & FewS \\ \hline
\multirow{7}{*}{Ger.} & BPE & No & 32 & 28 & 42 & 39 & 33 & 50 & 28 & 23 & 37\\
 & Mix & No & 42 & 38 & 48 & 50 & 38 & 58 & 37 & 34 & 43\\ \cline{2-12}
& BPE & Annot & 47 & 40 & 61 & 61 & 52 & 80 & 35 & 39 & 47\\
& Mix & Annot & 66 & 65 & 68 & 83 & 81 & 88 & 51 & 47 & 56\\ \cline{2-12}
& BPE & add. Annot & 51 & 49 & 55 & 65 & 62 & 70 & 37 & 36 & 38\\
& Mix & add. Annot & 65 & 63 & 69 & 81 & 78 & 88 & 51 & 40 & 56\\ 
& Mix+Annot & add. Annot & 72 & 72 & 72 & 92 & 91 & 94 & 58 & 56 & 60\\ \hline
\multirow{5}{*}{Czech} & BPE & No & 34 & 25 & 53 & 44 & 32 & 67 & 33 & 24 & 51\\
 & BPE & Annot & 46 & 33 & 70 & 63 & 48 & 92 & 42 & 30 & 67 \\
 & Mix+Annot & Annot & 64 & 61 & 69 & 92 & 91 & 95 & 60 & 58 & 65 \\ \cline{2-12}
& BPE & add. Annot & 45 & 31 & 71 & 61 & 45 & 91 & 41 & 29 & 58 \\
 & Mix+Annot & add. Annot & 66 & 63 & 72 & 92 & 89 & 95 & 63 & 60 & 70 \\ \hline

\end{tabular}   
\end{center}
\caption{\label{EPPSTranslationAna} Translation Performance on the Europarl data set }
\end{table*}

In a second set of experiments, we evaluated the approach on the larger data set on two different language pairs. In addition to the two word representation from the last experiment (\textit{BPE} and \textit{Mix}), we also applied \textit{Mix+Annotate}, where we also represent all words within dictionary entries as characters as described in Section \ref{Representation}. Furthermore, we also investigate \textit{add. Annot}, where additional dictionary entries were used for more training examples. The results are shown in Table \ref{EPPSTranslationAna}.  

The overall picture for these experiments and the previous experiments is quite similar. For all three scenarios, the quality of the various systems on the official test sets is relatively similar, however the systems differ when looking especially at the accuracy of translating the dictionary entries. Only when combining one-shot learning with character-based representation, we are able to successfully translate the dictionary entries. Independent of the language pair and data size, we are able to achieve an accuracy of around 70\% and an accuracy of around 90\% when only looking at the lemmas only. Furthermore, the model performs as good in one-shot learning as in few-shot learning.

However, beside the evidence that the approach works on various language pairs and data sizes, the additional experiments give some more insights. First, although the data is larger, we do not see a difference between the models using additional annotation and the models using only the baseline annotation. So it seems to be sufficient to have around 1000 examples in order to learn to copy the suggestions from the source sentence.

Furthermore, although there are no longer clear improvements for character-based representation on the overall translation performance, also for this experiment with larger data size these representations are essential for the dictionary integration. This is highlighted by the improvements of using characters for all words in dictionary entries (\textit{Mix+Annot}) instead of only for rare words (\textit{Mix}).

\section{Related work}

In recent years, several different approaches to integrate additional data into neural machine translation have been suggested. If this is parallel data, fine-tuning on the additional, better matching data \citep{luong_stanford_2015,lavergne_limsis_2011} is often successful. If the additional data is provided in other forms, different techniques have been investigated.

For human feedback, \citet{turchi_continuous_2017} suggested to use fine-tuning on the human generated post edits. \citet{pham_towards_2018} used phrase pairs extracted by statistical machine translation to annotate translations of rare phrases. In the similar scenario \citet{li_learning_2019} used a neural network to store the external phrase pairs.

Even more work has been done to integrate dictionaries into neural machine translation. A first work by \citet{arthur_incorporating_2016} used the additional dictionary to influence the softmax probabilities of the neural machine translation. Another possibility is to include the dictionary as an additional knowledge source during training using posterior regularization \citep{zhang_prior_2017}. A different approach is chosen by \citet{zhang_bridging_2016} using the dictionary as additional training sentences or generating synthetic sentences. In contrast to this work, these do not allow the integration of new words after training the NMT system. 

Several authors investigate the integration of the dictionary as an additional constraint during the coding process \citep{chatterjee_guiding_2017,hokamp_lexically_2017,hasler_neural_2018}. This leads to a larger complexity in decoding that has been addressed by \citet{post_fast_2018}. However, the dictionary is typically a hard constraint which makes it difficult to learn words forms that do not occur in the dictionary.

Most similar to this work is the approach by \citet{dinu_training_2019}, which like this work and \citet{pham_towards_2018} annotates the source sentence with possible translations. They showed that state-of-the-art models no longer need architecture changes, but can directly learn to copy form the source sentences. In this work, we additionally focus on generating new word morphological forms not occurring in the dictionary. We investigated different word representations and analysed their influence on the ability to copy the dictionary entries.

\section{Conclusion}

By introducing the new continuous learning test set using a different train-test split for existing corpora we could highlight the challenges of state-of-the-art neural machine translation systems. While they achieve very good performance, they are still challenged by new emerging terms. The baseline system was only able to correctly translate 20 to 30 percent of these phrases. 

Our integration of bilingual dictionaries into the systems improves the translation performance to correctly translate the words by up to 70\%. In 90\% of the cases at least the lemma of the word is predicted correctly. Furthermore, in this case, we see no difference in accuracy between words only seen in the dictionary and words also seen a few times in the parallel data. However this is only possible by modelling both: enabling the model to perform one-shot learning and modeling the different morphological forms of the rare phrases. The first one is addressed by annotating the source sentence with dictionary translation while the second one is addressed by using character-based models. By combining character-based and word-based representations we are able to model the different morphological variants of a word as well as enabling the system for fast training.

As mentioned before, this work concentrates on the morphological variants of the dictionary entries and ignores ambiguities due to different possible translation. In the future, we intend to address this by including word sense disambiguation into the translation process.

\bibliography{anthology,references}
\bibliographystyle{acl_natbib}

\end{document}